\documentclass[11pt,letterpaper]{article}
\usepackage{emnlp2016}
\usepackage{times}
\usepackage{latexsym}
\usepackage{tablefootnote}

\emnlpfinalcopy

\def\emnlppaperid{***}

\usepackage{amsmath}
\usepackage{mathabx}
\usepackage{mathtools}

\usepackage[bookmarks=false]{hyperref}

\usepackage{booktabs}
\usepackage{multirow}												
\usepackage{array}
\usepackage{fix-cm}

\usepackage{threeparttable}

\title{Compositional Sentence Representation \\* from Character within Large Context Text}

\author{Geonmin Kim, Hwaran Lee, Jisu Choi, Soo-young Lee\\
  {Korea Advanced Institute of Science and Technology, Daejeon, South Korea} \\
  {\tt $\{$gmkim90, hwaran.lee, jisu916, sy-lee$\}$@kaist.ac.kr} }

\date{}

\begin{document}

\maketitle{}

\begin{abstract}
This paper describes a Hierarchical Composition Recurrent Network (HCRN) consisting of a 3-level hierarchy of compositional models: character, word and sentence. This model is designed to overcome two problems of representing a sentence on the basis of a constituent word sequence. The first is a data sparsity problem when estimating rare words, and the other is no usage of inter-sentence dependency. In the HCRN, word representations are built from characters, thus resolving the data-sparsity problem, and inter-sentence dependency is embedded into sentence representation at the level of sentence composition. We propose a hierarchy-wise language learning scheme in order to alleviate optimization difficulties when training deep hierarchical recurrent networks in end-to-end fashion. The HCRN was quantitatively and qualitatively evaluated on a dialogue act classification task. Especially, sentence representations with inter-sentence dependency are able to capture both implicit and explicit semantics of sentences, significantly improving performance. In the end, the HCRN achieved state-of-the-art performance with a test error rate of 22.7$\%$ for dialogue act classification on the SWBD-DAMSL database.
\end{abstract}

\section{Introduction}
\label{sec: Introduction}
Sentence representations are usually built from representations of constituent word sequences by a 
compositional word model. Many compositional word models based on neural networks have been 
proposed, and have been used for sentence classification \cite{socher10,kim14}  or
generation \cite{sutskever14}  tasks. However, learning to represent a sentence on the basis of constituent word sequences faces two difficulties.First, vector representation of word, word embedding, vary independently from other words. Thus, estimation of rare words suffer from the data-sparsity problem in which there are insufficient data samples to learn embedding of them. Poorly estimated of rare words embedding can cause sentence representations of inferior quality. Second, conventional sentence representation does not take into account the dependency of the meaning of one sentence on the meanings of other sentences. This inter-sentence dependency is especially evident in large context text such as in documents and in dialogues. Without accounting for this inter-sentence dependency, model capture only superficial meanings of sentences, and cannot capture implicit aspects of sentences such as intention which often requires linguistic context to understand.

\begin{figure*}[t]
 \centering
  \includegraphics[width=16cm] {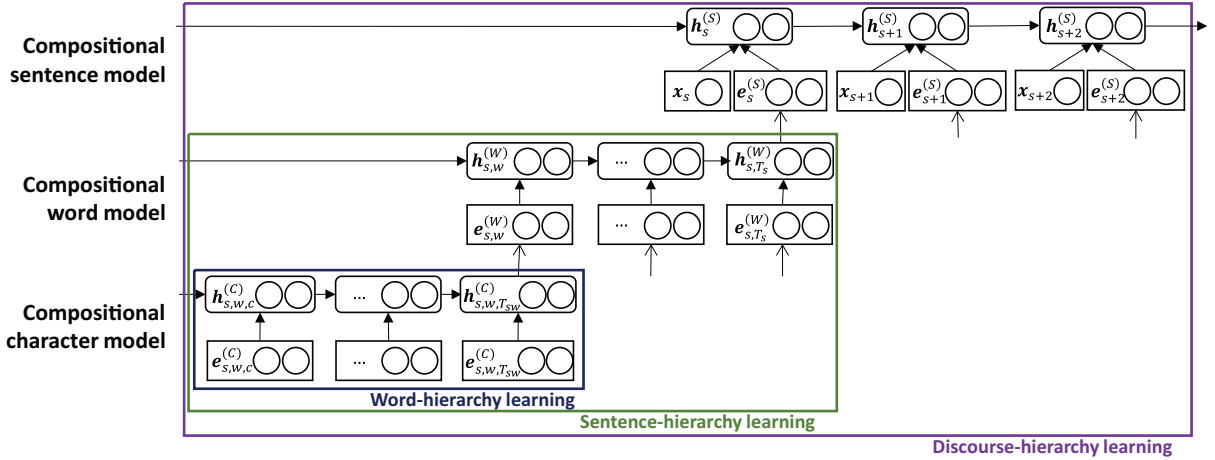}
   \caption{ Illustration of the Hierarchical Composition Recurrent Network. The thick arrows indicate transformations described in equations (3)-(6), the thin arrows lines indicate identity transformation. For simplicity, each level is shown with one layer.}
 \label{HCRN}
\end{figure*}

In this paper, we propose a Hierarchical Composition Recurrent Network (HCRN), which consists of a hierarchy of 3 levels of compositional models: character, word and sentence. Sequences at each
level are composed by a Recurrent Neural Network (RNN) which has shown good performance on various sequence modeling tasks. In the HCRN, output of lower levels of the compositional model is fed into the higher levels. Sentence representation by the HCRN enjoys several advantages when compared to sentence representation by a single compositional word model. From the compositional character model, the word representation is built from characters by modeling morphological processes shared by different words. In this way, the data-sparsity problem with rare words is resolved. From the compositional sentence model, inter-sentence dependency can be embedded into sentence representation. Sentence representation with inter-sentence dependency is able to capture implicit intention as well as explicit semantics of given sentence.

Training the HCRN in an end-to-end has presents optimization difficulties since the network, a deep hierarchical recurrent network, may suffer from the vanishing gradient problem across different levels in the hierarchy. To alleviate this, hierarchy-wise language learning algorithm is proposed, and empirically shows that it improves the optimization of network. The hierarchy-wise learning algorithm trains the lower level network first, and then trains higher levels successively. 

The efficacy of the proposed method is verified on a spoken dialogue act classification task. The task is to classify communicative intentions of sentences in spoken dialogues. Compared to conventional sentence classification, this task presents two challenging problems. First, it requires that the model estimate representations of spoken words which often include rare and partial words. Second, understanding dialogue context is often required to clarify meanings of sentences with a given dialogue. 

The HCRN with hierarchy-wise learning algorithm achieves state-of-the-art performance on the SWBD-DAMSL database.
\section{Hierarchical Composition Recurrent Network}
\label{sec: Model}

Figure~\ref{HCRN} shows our proposed Hierarchical Composition Recurrent Network (HCRN). The HCRN consists of a hierarchy of RNNs with compositional character, compositional word, and compositional sentence levels. At each level, each sequence is encoded by the output state of hidden layer of RNN at the end of sequence, which is  commonly used in other research \cite{cho14,tang15}. In Figure~\ref{HCRN} and in this section, each level of RNN is assumed to have one layer for simplicity of notation and figure. The notation of well-known transformations are represented as follows: gating units such as LSTM or GRU are represented as $g$, and affine transformation and non-linearity as $r$.

Consider a dialogue $D$ which consists of sentence sequences $s_{1:T_D}$ and its associated label $t_{1:T_D}$. The compositional character model sequentially takes the $c$-th character embedding $\mathbf{e}^{(C)}_{s,w,c}$ of word $w$ in sentence $s$, and recurrently calculates the output state of hidden layer $\mathbf{h}^{(C)}_{s,w,c}$ to produce a word representation $\mathbf{e}^{(W)}_{s,w}$ as:
\begin{equation}
      \mathbf{h}^{(C)}_{s,w,c} = g(\mathbf{h}^{(C)}_{s,w,c-1} ,\mathbf{e}^{(C)}_{s,w,c})
\end{equation}
\begin{equation}
      \mathbf{e}^{(W)}_{s,w} = \mathbf{h}^{(C)}_{s,w,T_{sw}}
\end{equation}
where $T_{sw}$ is the length of character sequence of word $w$ in sentence $s$. 

Similarly, the compositional word model takes word sequence as input, and iteratively calculates the output state of hidden layer $\mathbf{h}^{(W)}_{s,w}$ to produce a representation of the $s$-th sentence $\mathbf{e}^{(S)}_{s}$ as:
\begin{equation}
      \mathbf{h}^{(W)}_{s,w} = g(\mathbf{h}^{(W)}_{s,w-1} ,\mathbf{e}^{(W)}_{s,w})
\end{equation}
\begin{equation}
      \mathbf{e}^{(S)}_{s} = \mathbf{h}^{(W)}_{s,T_s}
\end{equation}
where $T_{s}$ is the number of words in sentence $s$.

The compositional sentence model updates its hidden neuron activation $ \mathbf{h}^{(S)}_{s}$ in the same way as lower levels. Especially for dialogue, the compositional sentence model additionally takes the agent (or speaker) identity change vector $\mathbf{x}_{s}$, as an input. Agent identity, or at least agent identity change across neighboring sentences, is an important clue to understand the intended meaning of a sentence in a dialogue, as shown in previous research \cite{li16}.
\begin{equation}
      \mathbf{h}^{(S)}_{s} = g(\mathbf{h}^{(S)}_{s-1} ,\mathbf{e}^{(S)}_{s},\mathbf{x}_{s}),
\end{equation}
where agent identity change vector $\mathbf{x}_{s}$ is:
\begin{equation}
      \mathbf{x}_{s} =
      \begin{cases}
     (1,0)^T, & \text{if}\ a_s = a_{s-1} \\
     (0,1)^T, & \text{if}\ a_s \neq a_{s-1}
       \end{cases}
\end{equation}
where $a_s$ is the agent identity of $s$-th sentence.

The memory of the compositional sentence model $\textbf{h}^{(S)}_{s}$, which includes $s$-th sentence information as well as that of previous sentences in the dialogue, is fed into a multi-layer Perceptron for classification of its label.

One advantage of the HCRN is its ability to learn long character sequences. While conventional stacked RNNs have difficulty when dealing with very long sequences \cite{bengio94,hochreiter98}, the hierarchy of the HCRN deals with short sequences of types specific to each level so that vanishing gradient problems during back-propagation through time are relatively insignificant. Each level of the HCRN uses a different speed of dynamics during sequence processing, so that the model can learn both short-range and long-range dependencies in large text samples. (See the preliminary experiment in supplementary materials of section A for details).

The following abbreviations are used for the rest of this paper: the compositional
character model ($CC$), the compositional word model ($CW$), the compositional sentence model ($CS$), and the multi layer perceptron ($MLP$).

\section{Hierarchy-wise Language Learning}
\label{sec: Lang_Hier_learn}
In order to alleviate optimization difficulties occurring when entire hierarchies of RNNs are trained in an end-to-end fashion, hierarchy-wise language learning is proposed. In the hierarchy of composition models, the lower level composition network is trained first, gradually adding higher level composition networks after the lower level network is optimized for a given objective function. This approach is inspired by the unsupervised layer-wise pre-training algorithm in \cite{hinton06}, known to provide better initialization for subsequent supervised learning.

\subsection{Unsupervised Word-hierarchy learning}
\label{ssec:Word-hierarchy learning}
\begin{figure}[t]
 \centering
  \includegraphics[width=7.5cm]{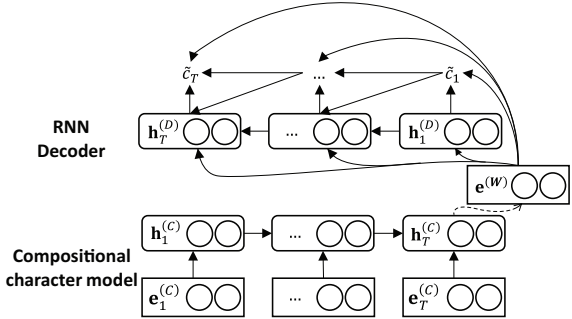}
   \caption{Illustration of the architecture used for word-hierarchy learning. Training objective forces the network to reconstruct exactly the same input character sequences.}
 \label{word_hierarchy}
\end{figure}

To pre-train the $CC$, we adopt a pre-training scheme by \cite{srivastava15}, while following the architecture of RNN Encoder-Decoder in \cite{cho14}. Figure~\ref{word_hierarchy} shows the RNN Encoder-Decoder architecture used for this learning, and parameters of the $CC$ can be obtained from the RNN Encoder. 

In this architecture, the representation of word $\textbf{e}^{(W)}$ which consists of characters $c_{1:T_W}$ is built by the $CC$. This  representation is then fed into the RNN decoder. The $CC$ and RNN Decoder are jointly trained so that their output sequence ${\tilde{c}_{1:T_W}}$ becomes exactly the same as the input character sequence $c_{1:T_W}$ by minimizing the negative log likelihood. This phase helps the $CC$ learns \textit{how to spell} words as character sequences, reducing the burden of learning morphological process of words on subsequent learning in higher levels of the hierarchy.

\subsection{Supervised Sentence-hierarchy and Discourse-hierarchy learning}
\label{ssec:Sentence,Discourse-hierarchy learning}
Next, sentence-hierarchy learning and discourse-hierarchy learning proceed in a supervised way with given sentence labels. In sentence-hierarchy learning, the model is trained to classify the label of single sentences independently from other sentences in a context. In discourse-hierarchy learning, the model is trained to classify the label across multiple sentences. An $MLP$ is stacked on top of sentence representation $\mathbf{e}_{s}^{(S)}$ (for sentence-hierarchy learning), or $\mathbf{h}_{s}^{(S)}$ (for discourse-hierarchy learning) to predict class labels.

A randomly initialized $CW$ is stacked on top of the $CC$ pre-trained during word-hierarchy learning, and a randomly initialized $CS$ is stacked on top of the pre-trained $CC$ and $CW$ resulting from sentence-hierarchy learning. The parameters of the pre-trained lower level models are excluded from the training during the first few epochs in order to prevent the lower level pre-trained models from changing too much due to large error signals generated by the randomly initialized layer. This method is also employed in \cite{oquab14} when adding a new layer on top of pre-trained layers for transfer learning. In section 4.5, the hierarchy-wise learning will be empirically shown to alleviates optimization difficulties in training the HCRN compared to end-to-end learning.

\section{Experiments}
\label{sec: Experiments}

\subsection{Task and Dataset}
\label{sec: Task and Dataset}

The HCRN was tested on a spoken dialogue act classification task. The dialogue act (DA) is the communicative intention of a speaker for each sentence. Prediction of the DA can be further used as an input to modules in dialogue systems such as dialogue manager. We chose the SWBD-DAMSL database\footnote{The dataset is available at \url{https://web.stanford} \newline \url{.edu/~jurafsky/swb1_dialogact_annot.tar.gz} }, which is a subset of the Switchboard-I (LDC97S62) corpus annotated with DA for each sentence. The Switchboard-I consists of phone conversations between strangers. The SWBD-DAMSL has 1155 dialogues of 70 pre-defined topics, 0.22M sentences, and 1.4M word tokens. There are several dialogue act tagsets available depending on the purpose of application. We chose
the 42-class tagset from DAMSL, which is widely used to analyze dialogue acts of phone
conversations. (See supplementary material section A for the complete class list.)

The number of elements in the character dictionary is 31 including 26 letters, - (indicating a partial word), \rq (indicating possessive case), . (indicating abbreviation),  $\textless$noise$\textgreater$ (indicating non-verbal sound) and  $\textless$unk$\textgreater$ (indicating unknown words) for all other characters. We follow the train/test set division in \cite{stolcke00}: 1115/19 dialogues, respectively. Validation data includes 19 dialogues chosen from training data. After pre-processing of the corpus, the number of sentences in the train/test/validation sets are 197370, 4190 and 3315 respectively. (See supplementary material section C for pre-processing details.) Table~\ref{hierarchy_statistic} shows the sequence length statistics. Note that a sentence becomes a much longer sequence when represented by characters (37.92) compared to words (8.28).


\setlength{\tabcolsep}{5pt}

\begin{table}[htbp]
  \centering
  \caption{Summary of the average sequence length for each level. C, W, S, and D indicate character, word, sentence and dialogue levels of processing, respectively. \#A/B means the average number of A per B.}
  {\small
    \begin{tabular} {c|ccc||c}
    \hline
    Hierarchy & \#C/W & \#W/S & \#S/D & \#C/S \\
    \hline
    Mean  & 8.19  & 8.28  & 161.26 & 37.92 \\
    Stddev & 2.52  & 8.11  & 67.32 & 39.72 \\
    \hline
    \end{tabular}%
    }
  \label{hierarchy_statistic}%
\end{table}%

\begin{table}[htbp]
  \centering
  \caption{Size of compositional model at each level, represented by ($\#$ layers) $\times$ ($\#$ memory in each layer). Note that the complexity of the model increases as the level of hierarchy increases, following the assumption that the complexity of composition increases as the level of language increases.}
  {\footnotesize
    \begin{tabular}{cccc}
    \toprule
          &CC&CW&CS\\
    \midrule
    Small & $1\times 64$ & $2\times 128$ & $2\times 256$ \\
    Large & $2\times 128$ & $3\times 256$ & $3\times 512$ \\
    \bottomrule
    \end{tabular}%
    }
  \label{size_of_hierarchy}%
\end{table}%

\subsection{Common settings}
\label{ssec:Common settings}

We employ the Gated Recurrent Unit (GRU) as a basic unit of the RNN because several studies have shown that the GRU performs similar with an LSTM, while requiring that fewer parameters be tuned \cite{chung14,jozefowicz15}. The configuration of the HCRN is represented by the hierarchy of the compositional model and its size, $CC_{size}-CW_{size}-CS_{size}$, where the size is represented by the number of layers and the number of hidden units in each layer. We tested two different sizes of compositional model at each level, as shown in Table~\ref{size_of_hierarchy}.

In all supervised learning, the classifier consists of MLP with 3-layers of affine transformation and non-linearity. The non-linearity of the first 2 layer is Rectified Linear Unit (ReLU), and that of last layer is softmax. The number of hidden neurons used in MLP is 128. A common hyperparameter setting is used in all experiments. All weights are initialized from a uniform distribution within [-0.1, 0.1] except pre-trained weights. We optimized all networks with adadelta \cite{zeiler12} with decay rate ($\rho$) 0.9 and constant ($\epsilon$) $10^{-6}$ which show much faster convergence than stochastic gradient descent with momentum 0.9, and employed a gradient clipping strategy adopted from \cite{pascanu13} with clipping threshold 5. Early stopping based on validation loss was used to prevent overfitting. 

\subsection{Unsupervised Word-hierarchy learning}
\label{sec: Unsupervised Word-hierarchy learning}
During word-hierarchy learning, $CC$ is jointly trained with the RNN Decoder to reconstruct input character sequences. The number of all unique words in the training set is 19353. The \textit{end of word} token is appended at every end of character sequence. The parameters are updated after processing a minibatch of 10 words. The chosen character embedding dimension is 15. Learning is terminated if validation loss fails to decrease by 0.1$\%$ for three consecutive epochs. 


\setlength{\tabcolsep}{1pt}

\begin{table}[]
\centering
\caption{Reconstruction performance of the RNN Encoder-Decoder on words in the vocabulary and out-of-vocabulary (OOV). The length column presents the mean and the standard deviation (in parentheses) of character length of words for which complete  reconstruction failed.}
\label{word_hierarchy_result}
{\footnotesize
\begin{tabular}{*{7}{c}}
\hline
\multirow{2}{*}{Model} & \multicolumn{3}{c}{In Vocabulary} & \multicolumn{3}{c}{Out of Vocabulary} \\ \cline{2-7} 
                       & {\begin{tabular}[c]{@{}c@{}}CPER\\ ($\%$) \end{tabular}}  & {\begin{tabular}[c]{@{}c@{}}WRFR\\ ($\%$) \end{tabular}}  & Length  & {\begin{tabular}[c]{@{}c@{}}CPER\\ ($\%$) \end{tabular}}  & {\begin{tabular}[c]{@{}c@{}}WRFR\\ ($\%$) \end{tabular}} & Length    \\ \hline
$CC_{1\times 64}$      & 0.39  & 2.25  & 13.1(2.6)  & 2.06  & 9.17 & 12.3(2.2) \\
$CC_{2\times 128}$     & 0     & 0     & -          & 1.21  & 5.28 & 12.7(2.4) \\ \hline
\end{tabular}
}
\end{table}

Pre-training performance itself is evaluated by sequence reconstruction ability. For reconstruction, the RNN Decoder generates character sequences from the encoding $\mathbf{e}^{(W)}$ generated by $CC$. Generation is performed based on greedy sampling at each time step. The performance is evaluated on two measures: Character Prediction Error Rate (CPER) and Word Reconstruction Fail Rate (WRFR). CPER measures the ratio of incorrectly predicted characters to the reconstructed sequence. WRFR is the ratio of words where complete reconstruction fails out of the total words in test set.

 The reconstruction performance of the RNN Encoder-Decoder on words both in vocabulary and out-of-vocabulary (OOV) is summarized in Table~\ref{word_hierarchy_result}.  Overall, the model almost perfectly reconstructs character sequences of training data, and even generalizes well for the unseen words. The large size model outperforms the small size model. Almost all cases in which reconstruction failed involved sequences longer than 12 characters on average.

\subsection{Supervised Sentence-hierarchy learning}
\label{sec: Supervised Sentence-hierarchy learning}
During sentence-hierarchy learning, the parameters are updated after processing a minibatch of 64 sentences.


\setlength{\tabcolsep}{3pt}

\begin{table*}[htbp]
  \centering
  \caption{Comparison of word representation built from CC and non-compositional word embedding. The nearest 3 words by Euclidean distance are retrieved for given target word.}
    {\small
    \begin{tabular}{ccccccccc}    
    \toprule
    \multirow{2}[2]{*}{\begin{tabular}[c]{@{}c@{}}Word representation\\  method\end{tabular}} & \multicolumn{2}{c}{Unigram counts $\geqq$ 5} &       & \multicolumn{2}{c}{1 $\leqq$ Unigram counts $\leqq$ 4} &       & \multicolumn{2}{c}{Out of Vocabulary (OOV)} \\ \cline{2-9}
    
          & uh-huh & really &       & emphasizing & probab- &       & environmentalism & seventy-eights \\
    \midrule
    \multirow{3}[2]{*}{\begin{tabular}[c]{@{}c@{}}Compositional \\ Character Model\end{tabular}} & uh-oh & reall- &       & emphasize & probably & \multirow{3}[2]{*}{} & environmentalist & ninety-eight \\
          & huh-uh & real  &       & emphasis & probability &       & environmentals & seventeenth \\
          & um    & very  &       & surpassing & probable &       & environmental & twentiy-six \\ \hline
    \multirow{3}[2]{*}{\begin{tabular}[c]{@{}c@{}} Non-compositional \\ ($\tau_{c}=5$) \end{tabular}} & hmm   & believe & \multirow{3}[2]{*}{} & \multicolumn{2}{c}{\multirow{3}[2]{*}{-}} & \multirow{3}[2]{*}{} & \multicolumn{2}{c}{\multirow{3}[2]{*}{-}} \\
          & helpful & very  &       & \multicolumn{2}{c}{} &       & \multicolumn{2}{c}{} \\
          & yeah  & frankly &       & \multicolumn{2}{c}{} &       & \multicolumn{2}{c}{} \\
    \bottomrule
    \end{tabular}%
    }
  \label{word_nn}%
\end{table*}%

\begin{figure}[t]
 \centering
  \includegraphics[width=7.5cm]{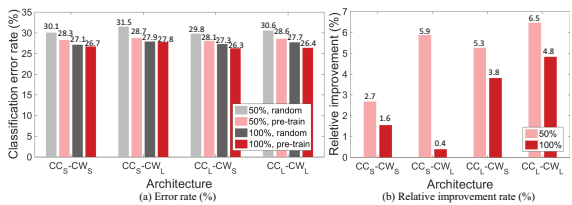}
   \caption{ Result to show the quality of our pre-trained CC for initialization on sentence-hierarchy learning. (a) Test error rate ($\%$) (b) Relative error improving rate of the pre-trained architecture with respect to the randomly initialized architecture. At each level, size is represented as either small (S) or large (L) in Table~\ref{size_of_hierarchy}.}
 \label{effect_of_pretrain}
\end{figure}

\textbf{Initialization of CC: Random VS. Pre-trained}
\label{sec: Effects of word-hierarchy pre-training}
The test set classification error rate of sentence-hierarchy learning with and without the pre-trained $CC$ are compared to evaluate how the pre-trained $CC$ provides useful initialization for sentence-hierarchy learning. With the pre-trained $CC$, at first, the parameters of the $CC$ are frozen, and the $CW$ and $MLP$ are trained for 1 epoch\footnote{The number of epochs to freeze the pre-trained model is chosen as the best parameter from preliminary experiments on the validation set.}. After that, the whole architecture consisting of the $CC$, $CW$, and $MLP$ is jointly trained. Evaluation was performed on architectures with different $CC$ and $CW$ sizes (see Table~\ref{size_of_hierarchy}). 
In addition, pre-training on two different training dataset sizes (50 $\%$  and 100 $\%$ ) are compared. The results are shown in Figure~\ref{effect_of_pretrain}. Pre-training consistently reduces the test error rate on the various architectures as shown in Fig. 3(a). Moreover, as shown in Fig. 3(b), improvement by pre-training is significant especially when fewer training data are available, where the model is liable to overfit to a given small number of data.

\textbf{CC VS. non-compositional word embedding}
\label{sec: Quality of word representation built by CC}
We compare two different methods to build word representation in this section: $CC$ and non-compositional word embedding. Note that pre-trained $CC$ are not directly compared to other widely used pre-trained word embedding such as Word2Vec \cite{mikolov13} since pre-training of the $CC$ aims at the model learning underlying morphological structures of words rather than learning the semantic/syntactic similarities between different words. Therefore, for a fair comparison we randomly initialized both models rather than employing pre-trained word embedding. 

For the non-compositional word embedding method, we set two different cutoff frequencies: $\tau_{c}=5$ (6294 words), $\tau_{c}=2$ (11746 words). The dimensions of the non-compositional word embedding are chosen as 64 and 128, which is the same dimensions produced by sizes of $CC$ in Table~\ref{size_of_hierarchy}.

\begin{figure}[t]
 \centering
  \includegraphics[width=7.5cm]{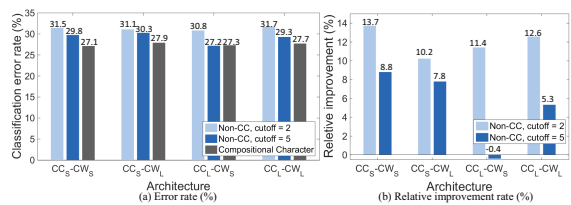}
   \caption{Result to show the quality of word representation built by CC. (a) Test error rate ($\%$) (b) Relative error improvement rate of the CC with respect to non-compositional word representation with two different cutoff frequencies. At each level, size is represented as either small (S) or large (L) listed in Table~\ref{size_of_hierarchy}.}
 \label{effect_of_compchar}
\end{figure}

Figure~\ref{effect_of_compchar} shows comparison of test error rates on above settings. Non-compositional word embedding with the high cutoff setting ($\tau_{c}=5$) outperforms the low cutoff setting ($\tau_{c}=2$). This is because the data sparsity problem during estimation of rare words is more severe for the model with the lower cutoff setting. Compared to the non-compositional method, $CC$ outperforms or is on par, with fewer parameters since the number of parameters for non-compositional word embedding scales with the number of words in a dictionary. 

Table~\ref{word_nn} shows the 3-nearest neighbors of word representation built by two different methods: $CC$ and non-compositional representation. For each method, the model with the best test accuracy is chosen. For word representation built by $CC$, retrieved nearest words usually have similar analogy with similar meaning. Moreover, rare words and OOVs such as partial words can be mapped to semantically similar words, which are usually estimated by a single $\textless$unk$\textgreater$ in non-compositional word representation.

\subsection{Discourse-hierarchy supervised learning}
\label{sec: Discourse-hierarchy supervised learning}
During discourse-hierarchy learning, the $CS$ on top of the  $CC_{2\times 128}-CW_{2\times 128}$ is trained. The pre-trained model was chosen by selecting the model with the lowest validation error rate during sentence-hierarchy learning. For the first 5 epochs, the network is trained with the $CC$ and $CW$ frozen. Then, the whole network is jointly optimized. The model is updated after it processes a minibatch of 8 dialogues. We evaluate performance of discourse-hierarchy learning with the two different sizes of $CS$ listed in Table~\ref{size_of_hierarchy}.

\begin{figure}[t]
 \centering
 \includegraphics[width=7.5cm]{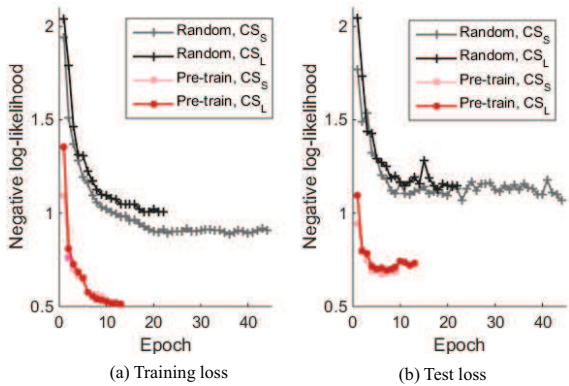}
   \caption{Learning curve on (a) training data and (b) test data. The objective function is converged to much lower value when the model employ initialization from the pre-trained model resulting from sentence-hierarchy learning.}
 \label{learning_curve}
\end{figure}

\textbf{Optimization difficulty of end-to-end learning}
\label{sec: Optimization difficulty of end-to-end learning}
To verify that hierarchy-wise learning actually alleviates optimization difficulies, we compared the objective function curves of discourse-hierarchy learning using two different model initializations: with the pre-trained model from sentence-hierarchy learning, and with random initialization (end-to-end learning). The learning curve in Figure~\ref{learning_curve} clearly shows that initializing with the pre-trained model significantly alleviates optimization difficulties.

\textbf{Effects of dialogue context on sentence representation}
\label{sec: Effects of dialogue context on sentence representation}
Table~\ref{effect_of_context} shows the test classification error rate of sentence-hierarchy learning and discourse-hierarchy learning. Compared to sentence-hierarchy learning, discourse-hierarchy learning improves performance significantly (up to 13.48$\%$ relative improvement of error).
 
To qualitatively analyze the improvement, we show examples of sentences on the test set for which prediction is improved by dialogue context. Analysis is done with model $CC_{2\times 128} - CW_{2\times 128} - CS_{2\times 256}$, which achieved the best test accuracy during discourse-hierarchy learning. Table~\ref{1improved_case} shows a dialogue segment example including 8 sentences, with the predicted labels for each sentence from both sentence-hierarchy and discourse-hierarchy learning. Highlighted sentences indicate cases where discourse-hierarchy learning correctly predicts while sentence-hierarchy learning fails to predict. For example, "yeah" in the 3rd sentence of the example can be interpreted as both Agreement and Backchannel, and an informed decision between the two is only possible when the dialogue context is available. This example demonstrates that sentence representation with dialogue context helps to distinguish confusing dialogue acts (see supplementary materials section D for further examples).


\setlength{\tabcolsep}{1pt}

\begin{table}[]
\centering
\caption{Test error rate of sentence hierarchy learning and discourse hierarchy learning. The relative improvement rate of error (Rel.) with respect to sentence-hierarchy learning are also provided.}
\label{effect_of_context}
\begin{tabular}{clcc}
\hline
{\footnotesize Hierarchy}                     & {\footnotesize Model}                                              & {\footnotesize Err ($\%$)}     & {\footnotesize Rel ($\%$)}           \\ \hline
 {\footnotesize Sentence}    & {\tiny $CC_{2\times 128}-CW_{2\times 128}$}   & {\footnotesize 26.27}          & -              \\ \hline
\multirow{2}{*}{{\footnotesize Discourse}} & {\tiny $CC_{2\times 128}-CW_{2\times 128}-CS_{2\times 256}$} & {\footnotesize \textbf{22.73}} & {\footnotesize \textbf{13.48}} \\ \cline{2-4} 
                          & {\tiny $CC_{2\times 128}-CW_{2\times 128}-CS_{3\times 512}$} & {\footnotesize 22.99}          & {\footnotesize 12.49}          \\ \hline
\end{tabular}
\end{table}
 
\begin{table*}[]
\centering
\caption{An example of dialogue segment containing 8 sentences. Predictions of label from model of sentence-hierarchy learning (without dialogue-context) and discourse-hierarchy learning (with dialogue-context) are provided along with true labels.}
\label{1improved_case}
{\footnotesize
\begin{tabular}{cccc}
\hline
\multicolumn{1}{c|}{Dialogue segment (with speaker)}                                         & \multicolumn{1}{c|}{True label}                                                     & \multicolumn{1}{c|}{\begin{tabular}[c]{@{}c@{}}Estimated label\\   without context \end{tabular}} & \multicolumn{1}{c}{\begin{tabular}[c]{@{}c@{}}Estimated label\\  with context \end{tabular}} \\ \hline
\multicolumn{1}{l|}{A: and uh quite honestly i just got so fed up with it}      & \multicolumn{1}{c|}{\multirow{2}{*}{Statement}}                         & \multicolumn{1}{c|}{\multirow{2}{*}{Statement}}                                          & \multicolumn{1}{c}{\multirow{2}{*}{Statement}}                                       \\ 
\multicolumn{1}{l|}{i just could not stand it any more}                          & \multicolumn{1}{c|}{}                                                               & \multicolumn{1}{c|}{}                                                                                & \multicolumn{1}{c}{}                                                                             \\ 
\multicolumn{1}{l|}{\textbf{B: is that right}}                                  & \multicolumn{1}{c|}{\textbf{Backchannel-question}}                                 & \multicolumn{1}{c|}{\textbf{Yes-No question}}                                                        & \multicolumn{1}{c}{\textbf{Backchannel-question}}                                               \\ 
\multicolumn{1}{l|}{\textbf{A: yeah}}                                           & \multicolumn{1}{c|}{\textbf{Agreement}}                                                 & \multicolumn{1}{c|}{\textbf{Backchannel}}                                                            & \multicolumn{1}{c}{\textbf{Agreement}}                                                               \\ 
\multicolumn{1}{l|}{\textbf{A: i mean this is the kind of thing you look at}}   & \multicolumn{1}{c|}{\textbf{Statement}}                                 & \multicolumn{1}{c|}{\textbf{\begin{tabular}[c]{@{}c@{}}Opinion\end{tabular}}}          & \multicolumn{1}{c}{\textbf{Statement}}                                               \\ 
\multicolumn{1}{l|}{B: yeah}                                                    & \multicolumn{1}{c|}{Backchannel}                                                    & \multicolumn{1}{c|}{Backchannel}                                                                     & \multicolumn{1}{c}{Backchannel}                                                                  \\ 
\multicolumn{1}{l|}{A: you sit there}                                           & \multicolumn{1}{c|}{Statement}                                          & \multicolumn{1}{c|}{Statement}                                                           & \multicolumn{1}{c}{Statement}                                                        \\ 
\multicolumn{1}{l|}{\textbf{A: and when you are writing up budgets you wonder}} & \multicolumn{1}{c|}{\multirow{2}{*}{\textbf{Statement}}}                & \multicolumn{1}{c|}{\multirow{2}{*}{\textbf{Wh-question}}}                                           & \multicolumn{1}{c}{\multirow{2}{*}{\textbf{Statement}}}                              \\ 
\multicolumn{1}{l|}{\textbf{okay how much money do we need}}                     & \multicolumn{1}{c|}{}                                                               & \multicolumn{1}{c|}{}                                                                                & \multicolumn{1}{c}{}                                                                              \\ \hline 
\end{tabular}
}
\end{table*} 

\textbf{Comparison with other methods}
\label{sec: Comparison with other methods}
 Several other methods for dialogue act classification are compared with our approach in Table~\ref{final_performance}. Our approach outperforms the other benchmarks, achieving 22.7$\%$ classification error rate on the test set.  Similar approaches employ a neural network based model which hierarchically composes sequences starting from word sequences \cite{nal13,ji16,serban15}. We conjecture that the improvement demonstrated by our model is due to two factors. First, our model build word representations from constituent characters and so suffers less from the data sparsity problem when confronted with rare words, and second the hierarchy-wise language learning method alleviates optimization difficulties of the deep hierarchical recurrent network. To the best our knowledge, our model achieves the state-of-the-art performance of dialogue act classification on the SWBD-DAMSL database. 


\begin{table}[htbp]
  \centering
  \caption{Performance comparison with other methods for dialogue act classification on SWBD-DAMSL.}
  {\footnotesize
    \begin{tabular}{lc}
    \toprule
    Method  & Test err. (\%) \\
    \midrule
    Class based LM + HMM \cite{stolcke00} & 29.0 \\
    RCNN \cite{nal13}  & 26.1 \\
    HCRN with word as basic unit\textsuperscript{$*$} & \multirow{2}[0]{*}{24.9} \\
    \cite{serban15}  \\
    Utterance feature + Tri-gram context & \multirow{2}[0]{*}{23.5} \\
    + Active learning + SVM \cite{bjorn11} &  \\
    Discourse model + RNNLM \cite{ji16} &  23.0\\
    \textbf{Discourse-hierarchy learning} & \textbf{22.7} \\
    \bottomrule
    \end{tabular}
  }  
  {\scriptsize
   \begin{tablenotes}
   \item[*] *This performance was evaluated by ourselves due to task difference. 
   \end{tablenotes}
  }
  \label{final_performance}
\end{table}

\section{Related works}
\label{Related work}

The difficulty for RNNs learning long-range dependencies within character sequences has been addressed in \cite{bojanowski16}. Hierarchical RNNs have been proposed as one possible solution. \cite{graves08,chan15} proposed sub-sampling sequences hierarchically to reduce sequence length at higher levels. \cite{koutnik14,chung15} proposed a RNN architecture wherein different layers learn at different speeds of dynamics. Compared with these models, the HCRN deals with shorter sequences at each level, and thereby the vanishing gradient problem is rendered relatively insignificant.

There are several recent studies on representing large context text hierarchically for sentence \cite{li15,serban15} and document classification \cite{tang15}. These approaches benefit from hierarchical representations which represents long sequences as a hierarchy of shorter sequences. However, the basic unit used in these approaches is the word, and models that begin at this level of representation open themselves to the data sparsity problem. This problem is somewhat resolved by building word representation from constituent sequences. Successful examples can be found in POS classification \cite{santos14} and language modeling \cite{botha14,ling15,kim15}.

\section{Conclusion}
\label{Conclusion}

In this paper, we introduced the Hierarchical Composition Recurrent Network (HCRN) model consisting of a 3-level hierarchy of compositional models: character, word and sentence. The inclusion of the compositional character model improves quality of word representation especially for rare and OOV words. Moreover, the embedding of inter-sentence dependency into sentence representation by the compositional sentence model significantly improves performance of dialogue act classification. This is because intentions which may remain ambiguous in single sentence samples are revealed in dialogue context, facilitating proper classification. The HCRN is trained in a hierarchy-wise language learning fashion, alleviating optimization difficulties with end-to-end training. In the end, the proposed HCRN using the hierarchy-wise learning algorithm achieves state-of-the-art performance with a test classification error rate of 22.7 $\%$ on the dialogue act classification task on the SWBD-DAMSL database.

Future work aims at the learning of hierarchy information of given sequential data without explicitly given hierarchy information. Another direction involves applying the HCRN to other tasks which might benefit from OOV-free sentence representation in large contexts such as document summarization.

\bibliography{Main}
\bibliographystyle{emnlp2016}


\newpage
\appendix
{
\onecolumn
{\LARGE
\textbf{Supplementary material}
}

\hfill
\hfill

\section{Effects of Hierarchical Composition}
\label{sec:Effects of Hierarchical Composition}

In order to learn character-level representation of large text such as sentence and dialogue, our model is hierarchically composed. As a preliminary experiment, we compared the performance of hierarchical (Fig 6-(a), sentence-hierarchy learning in paper) and non-hierarchical (Fig 6-(b), conventional stacked RNN) architectures on dialogue act classification tasks with the SWBD-DAMSL database. 

\hfill

\begin{figure}[!htbp]
 \centering
 \includegraphics[width=15cm]{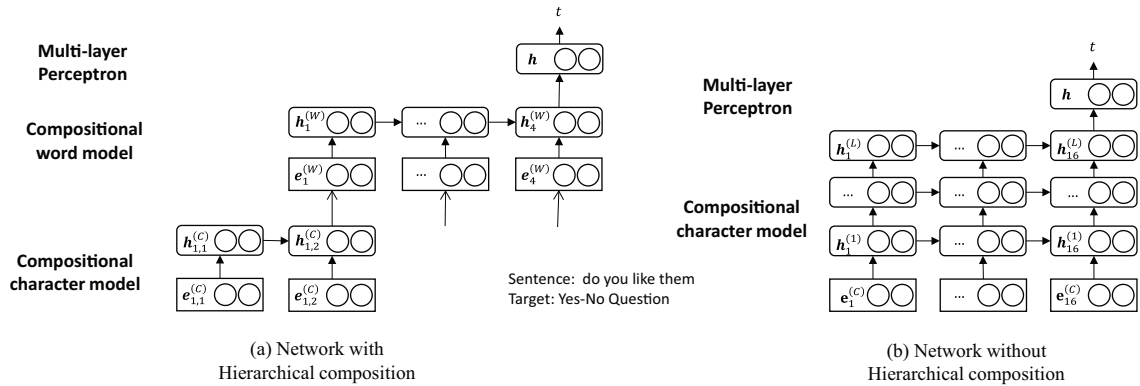}
   \caption{Comparison of network to learn sentence representation from character. (a) With hierarchical composition  (b) Without hierarchical composition(conventional stacked RNN)}
 \label{effect_of_hierarchical composition}
\end{figure}

\hfill

For fair comparison, hierarchical composition and non-hierarchical composition network have the same number of hidden units in each layer. For the non-hierarchical composition network, we add blank tokens in character sequences to indicate word boundaries. Table 8 shows that the hierarchical composition network outperforms the non-hierarchical composition network. It is because the network with hierarchical composition has a compositional word model (CW) which can be viewed as a specially designed compositional character model (CC), where composition is only allowed between hidden neurons at the ends of words. Therefore, the network with hierarchical composition has shorter sequences in each layer compared to the non-hierarchical model, where the vanishing gradient problem becomes relatively insignificant in each layer of RNN.

\hfill
\hfill

\begin{table}[!htbp]
\centering
\caption{Test error rate comparison of network with/without hierarchical composition. The relative improvement (Rel.) from non-hierarchical to hierarchical composition is also reported. ‘+’ indicates consecutive layers in a conventional stacked RNN.}
\label{Effect of Hierarchical composition}
{\normalsize
\begin{tabular}{|c|c|c|c|c|}
\hline
\multicolumn{2}{|c|}{Non-hierarchical composition}                                             & \multicolumn{3}{c|}{Hierarchical composition}                                                                                                                         \\ \hline
\multirow{2}{*}{Model} & \multirow{2}{*}{\begin{tabular}[c]{@{}c@{}}Error\\ (\%)\end{tabular}} & \multirow{2}{*}{Model} & \multirow{2}{*}{\begin{tabular}[c]{@{}c@{}}Error\\ (\%)\end{tabular}} & \multirow{2}{*}{\begin{tabular}[c]{@{}c@{}}Rel.\\ (\%)\end{tabular}} \\
                       &                                                                       &                        &                                                                       &                                                                      \\ \hline
$RNN_{1L,64H + 2L,128H}$                     & 35.80                                                                 & $CC_{1L,64H} - CW{2L,128H}$                     & 27.13                                                                 & 24.22                                                                \\ \hline
$RNN_{1L,64H + 3L,256H}$                       & 36.38                                                                 & $CC_{1L,64H} - CW{3L,256H}$                     & 27.81                                                                 & 23.56                                                                \\ \hline
\end{tabular}
}
\end{table}

\pagebreak

\section{Complete list of class names in SWBD-DAMSL}
\label{sec:Class list}

\begin{table}[!htbp]
  \centering
  \caption{42-class tagset of dialogue act provided from SWBD-DAMSL. Classes are sorted from the most frequent to the least frequent, from top-left to bottom-right with column-major order}
    {\footnotesize 
    \begin{tabular}{ccc}
    \toprule
    Non-Opinion & Declarative question & Other answers \\
    Backchannel & Backchannel(question) & Opening \\
    Opinion & Quotation & Or clause \\
    Abandoned & Summarize & Dispreferred answer \\
    Agreement & Non-yes answer & 3rd party talk \\
    Appreciation & Action-directive & Offers \\
    Yes-No-Question & Completion & Self talk \\
    Non-verbal & Repeat phrase & Downplayer \\
    Yes answer & Open question & Accept part \\
    Closing & Rhetorical question & Tag question \\
    Wh-question & Hold before answer & Declartive question \\
    No answer & Reject & Apology \\
    Acknowledgment & Non-no answer & Thanking \\
    Hedge & Non-understand & Others \\
    \bottomrule
    \end{tabular}%
    }
  \label{42tagset}%
\end{table}%

\section{Detail of pre-processing}
\label{sec:Detail of pre-processing}

We pre-processed raw text according to several rules below. At first, all letters are converted into
lower-case. Disfluency tags and special punctuation marks such as (? ! ,) which cannot be produced
by a speech recognizer are removed. We also merged sentences with segment tags into previous
unfinished sentences. Segment tags indicate the interruption of one speaker by another. It is difficult even for human beings to predict tags of segmented sentences, because sentence segments often do not provide enough information for reliable ascription of its DA. Sentences which interrupt others are placed after combined sentences. This scheme is also used in (Webb et al., 2005; Milajevs and Purver, 2014).

\section{Supporting results showing discourse context improves sentence representation}
\label{sec:Supporting results for discourse context improves sentence representation}

\subsection{Comparison of sentence-hierarchy learning and discourse-hierarchy learning on class accuracy}

\begin{figure}[!htbp]
 \centering
 \includegraphics[width=15cm]{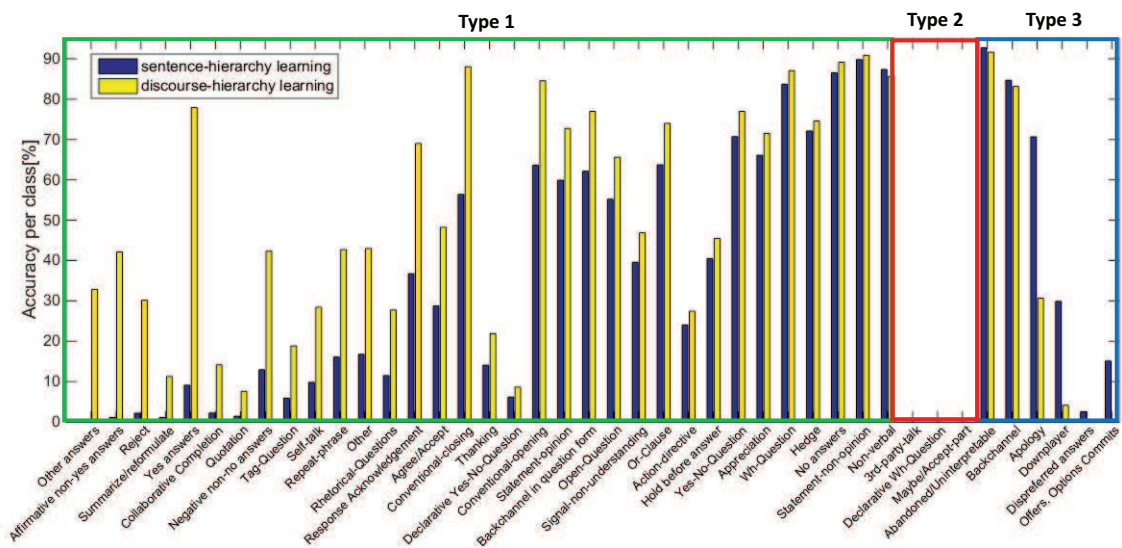}
   \caption{Class accuracy for sentence-hierarchy learning and discourse-hierarchy learning.}
 \label{effect_of_hierarchical composition}
\end{figure}

Figure 7 shows percent improvement when dialogue context is incorporated for each class label. Two models which achieves the best test set accuracy on sentence-hierarchy learning (without dialogue context) and dialogue-hierarchy learning (with dialogue context) are compared. Classes are sorted by descending order of relative improvement rate (from sentence-hierarchy to discourse-hierarchy). 33 out of 42 class improved with discourse-hierarchy learning (Type 1). Performance is degraded with discourse-hierarchy learning for 6 classes (Type 3). Also, there are 3 classes where both methods fail to predict. (Type 2)

\subsection{Improved cases}

\begin{table}[!htbp]
\centering
\caption{Examples of improved dialogue segment containing 8 sentences. More examples of Table 7.}
\label{improved_2cases}
{\footnotesize
\begin{tabular}{|c|c|c|c|}
\hline
\multirow{2}{*}{Text}                                                                                                                                                       & \multirow{2}{*}{True label}               & \multirow{2}{*}{\begin{tabular}[c]{@{}c@{}}Estimated label\\ without context\end{tabular}} & \multirow{2}{*}{\begin{tabular}[c]{@{}c@{}}Estimated label\\ with context\end{tabular}} \\
                                                                                                                                                                            &                                           &                                                                                            &                                                                                         \\ \hline
B : what d- what is that                                                                                                                                                    & Wh-Question                               & Wh-Question                                                                                & Wh-Question                                                                             \\
A : it's more uh $\textless$noise$\textgreater$                                                                                                                                 & Abandoned                                 & Abandoned                                                                                  & Abandoned                                                                               \\
A : i don't know how to explain it                                                                                                                                          & Hold before answer                        & Opinion                                                                                    & Opinion                                                                                 \\
A : kind of pop you know rock                                                                                                                                               & Statement                                 & Statement                                                                                  & Statement                                                                               \\
\textbf{B : rock}                                                                                                                                                           & \textbf{Repeat phrase}                    & \textbf{Statement}                                                                         & \textbf{Repeat phrase}                                                                  \\
\textbf{A : yeah}                                                                                                                                                           & \textbf{Agree}                            & \textbf{Backchannel}                                                                       & \textbf{Agree}                                                                          \\
\textbf{B : hard rock}                                                                                                                                                      & \textbf{Summarize}                        & \textbf{Statement}                                                                         & \textbf{Summarize}                                                                      \\
\textbf{A : well not hard $\textless$laughter$\textgreater$}                                                                                                                    & \textbf{Reject}                           & \textbf{Statement}                                                                         & \textbf{Reject}                                                                         \\ \hline
\multirow{2}{*}{\textbf{\begin{tabular}[c]{@{}c@{}}A : huh well are you going to paint the \\ outside of your house too\end{tabular}}}                                      & \multirow{2}{*}{\textbf{Yes-No-Question}} & \multirow{2}{*}{\textbf{Declarative question}}                                             & \multirow{2}{*}{\textbf{Yes-No Question}}                                               \\
                                                                                                                                                                            &                                           &                                                                                            &                                                                                         \\
B : well yeah                                                                                                                                                               & Yes answers                               & Yes answers                                                                                & Yes answers                                                                             \\
B : i think i am going to do it this spring actually                                                                                                                        & Statement                                 & Statement                                                                                  & Statement                                                                               \\
A : oh really                                                                                                                                                               & Backchannel-question                      & Backchannel-question                                                                       & Backchannel-question                                                                    \\
\textbf{B : yeah}                                                                                                                                                           & \textbf{Yes answers}                      & \textbf{Backchannel}                                                                       & \textbf{Yes answer}                                                                     \\
B : there are six houses                                                                                                                                                    & Statement                                 & Statement                                                                                  & Statement                                                                               \\
\multirow{2}{*}{\textbf{\begin{tabular}[c]{@{}c@{}}B : see the people that own the house they uh pay \\ for anything like that we do as far as the materials\end{tabular}}} & \multirow{2}{*}{\textbf{Statement}}       & \multirow{2}{*}{\textbf{Opinion}}                                                          & \multirow{2}{*}{\textbf{Statement}}                                                     \\
                                                                                                                                                                            &                                           &                                                                                            &                                                                                         \\
\multirow{2}{*}{\begin{tabular}[c]{@{}c@{}}A : there are three houses on this street the same \\ color of yellowout of six houses\end{tabular}}                             & \multirow{2}{*}{Statement}                & \multirow{2}{*}{Statement}                                                                 & \multirow{2}{*}{Statement}                                                              \\
                                                                                                                                                                            &                                           &                                                                                            &                                                                                         \\ \hline
\end{tabular}
}
\end{table}

}

\end{document}